\documentclass{article}

\usepackage[preprint]{neurips_2023}

\usepackage{graphicx}
\usepackage[utf8]{inputenc} 
\usepackage[T1]{fontenc}    
\usepackage{url}            
\usepackage{booktabs}       
\usepackage{amsfonts}       
\usepackage{nicefrac}       
\usepackage{microtype}      
\usepackage{xcolor}         
\usepackage[numbers]{natbib}
\usepackage{array}
\usepackage{wrapfig}
\usepackage{multirow}
\usepackage{enumitem}

\definecolor{redlinkcolor}{rgb}{0.79607843, 0.25098039, 0.25882353}
\definecolor{bluecitecolor}{rgb}{0,0.36,0.69}
\usepackage[colorlinks=true,linkcolor=redlinkcolor,citecolor=bluecitecolor,urlcolor=bluecitecolor]{hyperref}       
\title{Investigating Data Contamination for Pre-training Language Models}

\author{%
  Minhao Jiang$^{1}$, Ken Ziyu Liu$^{2}$, Ming Zhong$^{1}$, Rylan Schaeffer$^{2}$, \\
  \textbf{Siru Ouyang}$^{1}$\textbf{, Jiawei Han}$^{1}$\textbf{, Sanmi Koyejo}$^{2}$ \\
  $^{1}$University of Illinois Urbana-Champaign $^{2}$Stanford University\\
   \texttt{minhaoj2@illinois.edu}
}

\begin{document}

\maketitle

\begin{abstract}
Language models pre-trained on web-scale corpora demonstrate impressive capabilities on diverse downstream tasks. However, there is increasing concern whether such capabilities might arise from evaluation datasets being included in the pre-training corpus --- a phenomenon known as \textit{data contamination} --- in a manner that artificially increases performance. There has been little understanding of how this potential contamination might influence LMs' performance on downstream tasks. In this paper, we explore the impact of data contamination at the pre-training stage by pre-training a series of GPT-2 models \textit{from scratch}.
We highlight the effect of both text contamination (\textit{i.e.}\ input text of the evaluation samples) and ground-truth contamination (\textit{i.e.}\ the prompts asked on the input and the desired outputs) from evaluation data.
We also investigate the effects of repeating contamination for various downstream tasks 
Additionally, we examine the prevailing n-gram-based definitions of contamination within current LLM reports, pinpointing their limitations and inadequacy. Our findings offer new insights into data contamination's effects on language model capabilities and underscore the need for independent, comprehensive contamination assessments in LLM studies.
\end{abstract}

\section{Introduction} \label{sec::intro}

The performance of large language models (LLMs) has been attributed primarily to their immense size and the increasing scale of pre-training data from large text corpora~\citep{radford2019rewon, brown2020language, openai2023gpt4, chowdhery2022palm, anil2023palm, touvron2023llama, touvron2023llama2}. Nevertheless, a critical aspect that remains under-explored is the potential contamination of the pre-training corpus with evaluation data. This oversight presents challenges in accurately assessing the LLMs' capabilities among other scientific analyses of their behaviors. The importance of contamination analysis in the pre-training corpus has been recognized since pre-trained language models were first introduced~\cite{devlin-etal-2019-bert, radford2019rewon, chowdhery2022palm}; however, the lack of public access to most pre-training corpora today complicates efforts to comprehensively understand and identify the impact of contamination on a model's performance and behaviors.

Recent LLM reports~\citep{radford2019rewon,brown2020language,chowdhery2022palm,openai2023gpt4, touvron2023llama2, gunasekar2023textbooks} have investigated the contamination of evaluation data in the pre-training corpora from various perspectives. Some of these studies offer limited details on their contamination investigations, especially for closed-source models \cite{radford2019rewon, openai2023gpt4}. 
Others include attempts \cite{radford2019rewon, openai2023gpt4, brown2020language, chowdhery2022palm, touvron2023llama2} to investigate the data contamination on the \textbf{evaluation level}, where an evaluation dataset is \textit{post-hoc} categorized into contaminated and non-contaminated chunks based on a proposed contamination definition and the model is evaluated on them separately, to demonstrate that the model is insusceptible to data contamination if the model performs similarly on these chunks. However, this line of work has not adequately analyzed contamination on the \textbf{pre-training level}, where the pre-training corpus is deliberately altered to study the effects of contamination on evaluation.

Evaluation data can be leaked into the pre-training corpus in various formats. Predominantly, it is the textual component of the evaluation dataset (\textit{i.e.}\ input text). This aspect has been the primary focus of many existing studies (\textit{e.g.}~\cite{touvron2023llama2,chowdhery2022palm}). There are also many cases where the pre-training corpus might contain \textit{ground truth} information of the evaluation data. Here, we consider \textit{ground truth} of the evaluation samples to be their raw texts \textit{plus} the prompts on such texts and the corresponding answers.
Intuitively, contamination involving the ground truth may have different impacts on the models' performance than simple text contamination, but its effects have been under-explored.

Another recent line of work focuses on the detection of data contamination from the pre-training corpus through the lens of membership inference attacks~\cite{mahloujifar2021membership, mattern-etal-2023-membership, golchin2023time,oren2023proving, shi2023detecting}, which involves determining whether the given text is in the pre-training data of a black-box model.
While relevant, the detection of contamination does not necessarily offer a direct understanding of their effects during evaluation.
The recent works \cite{gunasekar2023textbooks, li2023textbooks} represent a step forward as they implement various methods, including embedding-based search and syntax-based similarity analysis, to both detect and filter contamination from pre-training corpora, although they primarily focus on code-based data.

This paper investigates the effects of contamination of pre-training data for language models via leakage of evaluation datasets. We pre-train \textit{from scratch} a series of GPT-2 models~\cite{radford2019rewon} and consider various mechanisms of contamination of evaluation data in the pre-training corpus. Specifically, we ask and answer three research questions:
\begin{enumerate}[leftmargin=*]
    \item \textbf{RQ1: How are language models affected by the deliberate addition of various forms of contamination on the pre-training corpus?}
    To answer this, we introduce intentional contamination (with and without the ground truth) into the pre-training corpus (\S\ref{sec::1}). We then pre-train GPT-2-small models \textit{from scratch} on these variously contaminated corpora to evaluate and compare their performance. We further extend the experiments with GPT-2-large models to evaluate the effects of data contamination on larger models (\S\ref{sec::large}).
    \item \textbf{RQ2: How do the number of repetitions of evaluation data in the pre-training corpus affect performance?} In practice, \textit{how often} a piece of evaluation data has appeared during pre-training and its ramifications are also unclear.
    We investigate this by injecting the evaluation data into the pre-training corpus multiple times and provide detailed empirical analyses (\S\ref{sec::2}).
    \item \textbf{RQ3: How effective are the n-gram-based contamination definitions used in recent LLM reports?} We systematically filter out different proportions of contaminated training documents, as described by these definitions, and pre-train the same model on these cleansed corpora (\S\ref{sec::3}). Additionally, we critically evaluate the methods used in current LLM reports for assessing data contamination at the evaluation level (\S\ref{sec::5}). These reports often posit that the models exhibit robustness against data contamination, and our discussion aims to elucidate the potential shortcomings of such claims.
\end{enumerate}

We evaluate our experiments on several commonly used public datasets to observe the performance differences quantitatively. Our analyses provide a new perspective on understanding data contamination in the pre-training of language models. The contributions are summarized as follows:
\begin{itemize}[leftmargin=*]
    \item We empirically investigate the effects of data contamination in the pre-training corpus due to evaluation data leakage in language models by pre-training language models from scratch to evaluate different mechanisms of data contamination.
    \item We identify the importance of considering the data contamination with ground truths from the evaluation dataset. Surprisingly, we observed that the effects of increasing the number of repetitions of contamination on the model performance can be U-shaped.
    \item We critically analyze the n-gram data contamination definitions from existing LLM reports and further compare the empirical results by filtering the pre-training data with these definitions. Our findings suggest that they are insufficient and inadequate to identify contamination.
\end{itemize}

\section{Contamination Definitions}
\label{sec::def}
Numerous studies on large language models (LLMs) have explored and investigated the concept of data contamination and demonstrated the robustness of these models against potential contamination in their evaluation datasets \cite{radford2019rewon, brown2020language, chowdhery2022palm, openai2023gpt4, touvron2023llama, touvron2023llama2, gunasekar2023textbooks}. Most definitions proposed in the existing studies are based on n-gram duplication between pre-training data and evaluation data. For instance, PaLM~\citep{chowdhery2022palm} divides the evaluation data into two categories---``clean'' and ``contaminated''---based on whether at least 70\% of all possible 8-grams in the evaluation sample were seen at least once in the pre-training corpus.
Llama 2~\citep{touvron2023llama2} provides a more fine-grained definition: a token is considered contaminated if it appears in any token n-gram longer than 10 tokens in both the evaluation sample and the training set, and the contamination percentage of an evaluation sample is defined to be the percentage of tokens contaminated; the evaluation data are then divided into 4 buckets---``Clean'', ``Not Clean'', ``Not Dirty'', and ``Dirty''---based on the contamination percentage of each evaluation sample.
While intuitive, these contamination definitions primarily revolve around n-gram or token overlaps, which only target direct duplications present in both training and evaluation datasets and might provide both high false positive rate (since many semantically different texts have overlaps) and false negative rate (since simple paraphrasing can evade detection~\cite{yang2023rethinking}).
Moreover, investigations relying on these definitions have predominantly centered on evaluation level analysis; in our work, we focus on pre-training level analysis as described in \S\ref{sec::intro}.

In our experiments, we follow PaLM~\citep{chowdhery2022palm} and Llama 2's~\citep{touvron2023llama2} definitions as well as a direct n-gram overlap detection strategy to investigate how the ``contamination'' under these definitions are different and how they affect model performance. As described in \S\ref{sec::intro}, contamination in the pre-training corpus can appear as either textual components from evaluation datasets or with ground truth information. Existing definitions tend to overlook the latter. Therefore, we explore two types of contamination when we introduce contamination to the pre-training corpus: (1) \textbf{text contamination}, where only the input texts of the evaluation samples are added to the pre-training corpus; and (2) \textbf{ground-truth contamination}, where the input texts, the prompts, and the labels/answers of the corresponding evaluation samples are added.
\section{Experimental Setup}

\subsection{Models, Data and Pre-Training}
\label{sec::pretraining}
The model architecture used in our main experiments is GPT-2-small~\citep{radford2019rewon} (124M parameters) with default hyperparameters. We use a relatively small architecture because pre-training from scratch is computationally expensive.
Following~\citep{korbak2023pretraining}, we construct a pre-training corpus by subsampling 1.95M documents from the Pile~\citep{gao2020pile} for a total of 3.3B tokens, which is compute-optimal based on Chinchilla scaling laws~\citep{hoffmann2022an}. We later extend our experiments to GPT-2-large (774M parameters) and 19.8B tokens from \texttt{pile-uncopyrighted} corpus\footnote{\url{https://huggingface.co/datasets/monology/pile-uncopyrighted}} (Sec. \ref{sec::large}), again following compute-optimal scaling laws. The detailed hyperparameters for all experiments are listed in Appendix \ref{sec::param}.

\subsection{Evaluation Datasets}

We focus our experiments on four natural language processing datasets to evaluate the performance of our pre-trained models: SST-2~\citep{2013sst2}, a sentiment analysis dataset; MMLU~\citep{hendrycks2021measuring}, a multi-task natural language understanding dataset; CNN And Daily News~\citep{nallapati-etal-2016-abstractive}, a text summarization dataset that was also evaluated in the GPT-2 report~\citep{radford2019rewon}; the Stanford Question Answering Dataset (SQuAD) dataset~\citep{rajpurkar2016squad}, which helps evaluating the reading comprehension abilities of the model. The detailed statistics of these datasets are listed in Table \ref{table::dataset}.
All datasets are accessed through HuggingFace\footnote{\url{https://huggingface.co/datasets/}}. We selected these easier and traditional benchmarks because our goal in the paper is to assess the differential impact of data contamination on GPT-2 models' performance, and the more difficult datasets are likely too challenging for GPT-2 series models.

\begin{table}[h]
    \centering
    \caption{\textbf{Evaluation Dataset Statistics.} The last column (\# of Samples) shows the number of evaluation examples corresponding to each label. 
    }
    \label{table::dataset}
    \resizebox{1.0\textwidth}{!}{
    \begin{tabular}{l|c|c|cr}
        \toprule
        \textbf{Dataset Name}& \textbf{Split} & \textbf{Label Space} & \textbf{\# of Samples} \\
        \midrule 
        SST-2 & train & positive, negative& 37,569 / 29,780 \\
        \midrule
        MMLU & all/test & A, B, C, D (57 Subjects) & 3,222 / 3,462 / 3,582 / 3,776 \\
        \midrule
        CNN And Daily Mail & 3.0.0/test & - & 11,490 \\
        \midrule
        SQuAD V1 & validation & - & 10,600  \\
        \bottomrule
    \end{tabular}
    }
\end{table}

For evaluation, we follow established processes. For the SST-2 dataset, due to the uncontrollability and instability of the generated results from GPT-2 models, we utilize prompting and the possible labels as hypotheses and ask the model to score each hypothesis and use the highest one as the prediction.
To circumvent prompt sensitivity~\cite{liang2022holistic}, we evaluate the accuracy scores based on 10 different prompts for each model. The details of the prompts and the corresponding performance are listed in Appendix \ref{sec::eval}. For MMLU, we utilize AllenAI's official MMLU implementations\footnote{\url{https://github.com/allenai/open-instruct}} \cite{wang2023far} to compute the accuracy across 57 different subjects.

For the text summarization task, we follow the original implementation reported in~\citep{radford2019rewon} for evaluation. We add the text \texttt{TL; DR: "} after the article to induce the summarization generation. We then ask the model to generate 150 tokens with top-$k$ random sampling with $k = 2$ and use the first 3 sentences of the generated tokens as the summary. We evaluate the generated summaries on the commonly used ROUGE-1, 2, L scores~\citep{lin-2004-rouge} and UniEval~\citep{zhong2022unified} to provide a multi-dimensional evaluation. For the question-answering evaluation on SQuAD, we employ the official implementation.\footnote{\url{https://rajpurkar.github.io/SQuAD-explorer}} In this setup, we allow the model to generate up to 15 tokens, and the first sentence of the generated output is taken as the answer. We subsequently report F1 scores for the generated answers, determined by the overlap of tokens between the model's response and the ground truth. We selected SQuAD V1 to mitigate potential biases introduced by the many no-answer questions in the V2 dataset.

\section{Experiments \& Analyses} \label{sec::experiments}
In this section, we present the experiment results to understand how data contamination affects the models' performance quantitatively. We conducted experiments with three variations of contamination, described as follows. For the main experiments, we pre-train the GPT-2-small model from scratch on the corpus to evaluate the performance:
\begin{itemize}[leftmargin=*]
    \item GPT-2-small$_{\textit{original}}$ is the model pre-trained on the original corpus described in \S\ref{sec::pretraining}.
    \item GPT-2-small$_{\textit{text}}$ is the text contamination version of the model. We only add the texts of the corresponding evaluation samples to the training data to ensure that all the texts in the evaluation dataset were 100\% contaminated in the pre-training corpus. For MMLU, we also include the texts from the answer choices of each question.
    \item GPT-2-small$_{\textit{gt}}$ is the ground-truth contamination variation of the model. On top of the text contamination, we add the same prompt used for evaluation and the ground truth (\textit{e.g.}\ labels) following the text for each dataset; that is, in the format as \textit{``text + prompt + ground truth''}. For SST-2, we randomly select one out of the 10 prompt templates for evaluation for each evaluation sample and insert it in the corpus as contamination. 
\end{itemize}
As baselines, we further evaluate all datasets on the public checkpoints for GPT-2-small, medium, and large variations to more directly compare the performance, where the pre-training data for the public checkpoints are unknown.

\subsection{Effects of Contamination on Evaluation Performance}
\label{sec::1}
To quantify the effects of data contamination and how the text and ground-truth contamination are different, we directly compare GPT-2$_{\textit{original / text / gt}}$ on each dataset in Table \ref{table::other} and \ref{table::cnn}.


\begin{table}[ht]
    \centering
    \caption{\textbf{Evaluation results on SST-2, MMLU, and SQuAD V1 datasets.} For three variations of models, the experiments are run 3 times, i.e., each pre-training was run under 3 different random seeds, and shown as mean$_{ std}$. Since only single checkpoints exist for the public baselines (GPT-2-small, GPT-2-medium, GPT-2-large), we have no way of computing variance over multiple training runs.
    }
    \label{table::other}
    \resizebox{0.65\textwidth}{!}{
    \begin{tabular}{l|c|c|c|cr}
        \toprule
        \textbf{Model} & \textbf{Parameters} & \textbf{SST-2} & \textbf{MMLU} & \textbf{SQuAD V1}\\
        \midrule
        & & Accuracy & Accuracy & F1 Scores\\
        \midrule
        GPT-2-small$_{\textit{original}}$ & 124M & 48.34$_{ 2.32}$ & 22.87$_{ 0.09}$ & 9.07$_{ 0.19}$\\
        GPT-2-small$_{\textit{text}}$ & 124M & 54.89$_{ 0.80}$ &  23.03$_{ 0.05}$& 9.78$_{ 0.12}$\\
        GPT-2-small$_{\textit{gt}}$ & 124M &51.02$_{ 0.35}$ &  23.13$_{ 0.09}$ & 11.45$_{ 0.58}$\\
        \midrule
        GPT-2-small & 124M & 52.06 &  23.0 & 15.09\\
        GPT-2-medium & 354M & 55.21 &  23.6 & 19.94\\
        GPT-2-large & 774M & 54.01 &  23.0 & 17.87\\
        \bottomrule
    \end{tabular}
    }
\end{table}
\begin{table}[h]
    \centering
    \caption{\textbf{Evaluation results on CNN And Daily Mail dataset.} Similarly, each experiment is run three times to report the mean/std, and only single checkpoints exist for public baselines.}
    \label{table::cnn}
    \resizebox{\textwidth}{!}{
    \begin{tabular}{l|c|ccccccccr}
        \toprule
        \textbf{Model}  & \multicolumn{8}{c}{\textbf{CNN And Daily Mail}} \\
        \midrule
        &  ROUGE-1 & ROUGE-2 & ROUGE-L & Coherence & Consistency & Fluency & Relevance & Overall\\
        \midrule
        GPT-2-small$_{\textit{original}}$  & 24.76$_{ 1.33}$ & 8.33$_{ 0.30}$ & 16.44$_{ 0.93}$ &  0.5382$_{ 0.045}$ & 0.6020$_{ 0.013}$ & 0.7513$_{ 0.035}$ & 0.4952$_{ 0.044}$ & 0.5968$_{ 0.010}$ \\
        GPT-2-small$_{\textit{text}}$   & 26.84$_{ 0.45}$ & 9.03$_{ 0.16}$ & 17.91$_{ 0.27}$ & 0.5137$_{ 0.016}$ & 0.6686$_{ 0.121}$ & 0.8225$_{ 0.009}$ & 0.4648$_{ 0.014}$ & 0.6174$_{ 0.008}$ \\
        GPT-2-small$_{\textit{gt}}$  & 28.80$_{ 0.08}$ & 10.65$_{ 0.08}$ & 19.49$_{ 0.04}$ & 0.6390$_{ 0.032}$ & 0.7471$_{ 0.012}$ & 0.8480$_{ 0.001}$ & 0.5644$_{ 0.001}$ & 0.6996$_{ 0.015}$ \\
        \midrule
        GPT-2-small  & 27.97 & 9.43 & 18.34 & 0.5725 & 0.6954 & 0.8703 & 0.5525 & 0.6727\\
        GPT-2-medium  & 29.71 & 10.52 & 19.49 & 0.6976 & 0.7998 & 0.8989 & 0.6793 & 0.7689\\
        GPT-2-large  & 29.97 & 10.92 & 19.77 & 0.7259 & 0.8253 & 0.8997 & 0.6942 & 0.7863\\
        \bottomrule
    \end{tabular}
    }
\end{table}

The experimental results from the two tables reveal the impact of data contamination on model performance across different datasets. The introduction of contamination, either in the text or ground truth, improves model performance compared to the original pre-trained GPT-2 model. Notably, while text contamination does show some improvement in evaluation metrics, the extent of this enhancement is relatively modest. This is particularly evident in the SQuAD and CNN datasets, where the coherence and relevance scores under text contamination are sometimes lower than those of the original model in the CNN dataset. Conversely, ground-truth contamination generally yields significant performance improvements. However, in the SST-2 dataset, ground-truth contamination does not outperform text contamination. We hypothesize that this is because text classification tasks predominantly depend on the model's comprehension of the input text, rendering evaluation prompts and ground truths less impactful. In fact, they might introduce noise, particularly given that the input texts in the dataset are generally short and that the model is sensitive to prompt formatting. For the MMLU dataset, it's evident that this task presents a significant challenge for GPT-2-small models, as indicated by the poor performance of both the public checkpoints and our pre-trained models. Despite this inherent difficulty, it is noteworthy that we can still observe the performance improvements with the introduction of both types of contamination. Overall, these findings suggest that while both types of contamination can enhance the performance of language models, ground-truth contamination has a more pronounced positive effect on model performance than text contamination in general cases, especially for tasks that require an understanding of the instructions from evaluation prompts, such as CNN and SQuAD datasets.

The improvement of ground-truth contamination is more pronounced for the CNN dataset, where ground-truth contamination can even improve the model to surpass the performance of public checkpoints and achieve similar performance with the GPT-2-medium model. The experiment results also indicate that fluency, as measured by the UniEval metric, is still lower than the public model checkpoints. We suspect that this observation is due to the smaller scale of training data, where fluency might be more closely related to the model's overall language abilities. We can also observe that there is still an obvious gap between our pre-trained model and the public OpenAI's checkpoints, which shows the importance of the scale of training data. 

Viewed together, Tables~\ref{table::other} and \ref{table::cnn} demonstrate the effects of data contamination on downstream evaluation tasks and, in particular, the effects of ground-truth contamination. 
The results highlight the need for methods that can identify and differentiate ground-truth contamination in future studies.

\subsection{Effects of Repeated Contamination Can Be U-Shaped}
\label{sec::2}
\begin{figure}[ht]
    \centering
    \includegraphics[width=\textwidth]{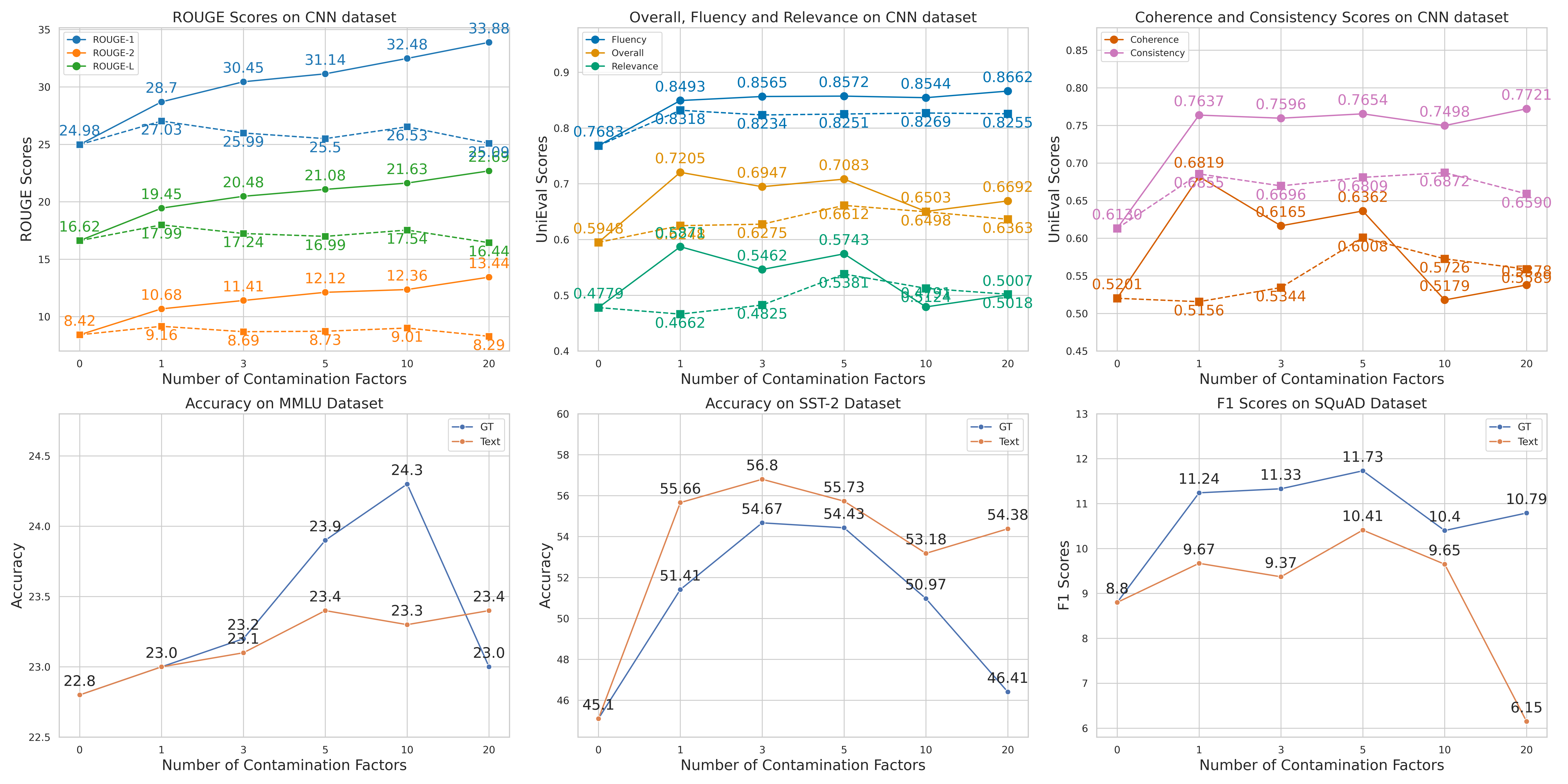}
    \caption{\textbf{Evaluation results for different contamination factors from 0 to 20 on each dataset}. Zero repetitions refer to models pre-trained on the original corpus. In the top three figures, the solid lines and the dotted lines show the ground-truth and text contamination results respectively.}
    \label{fig::repetition}
\end{figure}
We have already observed the effectiveness of data contamination in the previous section, where both the text and ground-truth contamination are only injected into the pre-training corpus once. However, in practice, some fractions of the evaluation datasets may appear in the pre-training corpus more than once given its immense scale. 
Therefore, in this section, we investigate the effects of \textit{repeated contamination} whereby the evaluation dataset is added to the pre-training corpus multiple times. We use the term \textbf{contamination factor} to denote the number of times the evaluation data appear in the pre-training corpus.
This analysis is designed to help us understand better how the repetitions of evaluation data for both text and ground-truth contamination, during pre-training might affect the performance. The results are shown in Figure \ref{fig::repetition}.

For SST-2, MMLU, and SQuAD datasets, we observed a distinct U-shaped performance trend in response to increasing contamination factors. Specifically, as the contamination factor increased, performance initially improved but started to decline when the factor reached around 10 repetitions. Notably, at 20 repetitions, performance in some instances dropped below the baseline level observed when there was no contamination. The results for the CNN dataset exhibited varying trends based on the evaluation metrics used. While the ROUGE scores steadily increased with higher contamination factors, the UniEval scores displayed a U-shaped curve similar to the other datasets, which also indicates a U-shaped general performance trend for the CNN dataset. Another observation to notice is that the fluency score also almost increases monotonically with the increase of contamination factor, which further indicates that fluency is more associated with the size of training data. The divergence in ROUGE scores is primarily attributed to the metrics' focus on the frequency of common subsequences and tokens. These elements are more likely to be repeated with increased data repetition, particularly in scenarios involving ground-truth contamination that repeats correct responses from the dataset. 

These findings suggest that while introducing contamination into a pre-training corpus can enhance model performance to a certain degree, over-repetition may lead to a decline in effectiveness.
We also note that this threshold for the number of repetitions can be related to the model size and corpus size, which requires more investigation in future works. This is an interesting result since many existing LLMs leveraged huge but \textit{unscrutinized} pre-training corpora that it is unclear: 1) how many times the evaluation data have appeared in the pre-training data, and 2) how the contamination has realistically affected evaluation performance.

On the other hand, we also observe that this U-shape curve for the contamination factor may not universally hold for all datasets and corpora, which we discuss in more detail in Appendix \ref{app::factor}.

\subsection{Effects of Removing Contamination from Pre-Training}
\label{sec::3}
\begin{figure}[ht]
    \centering
    \includegraphics[width=\textwidth]{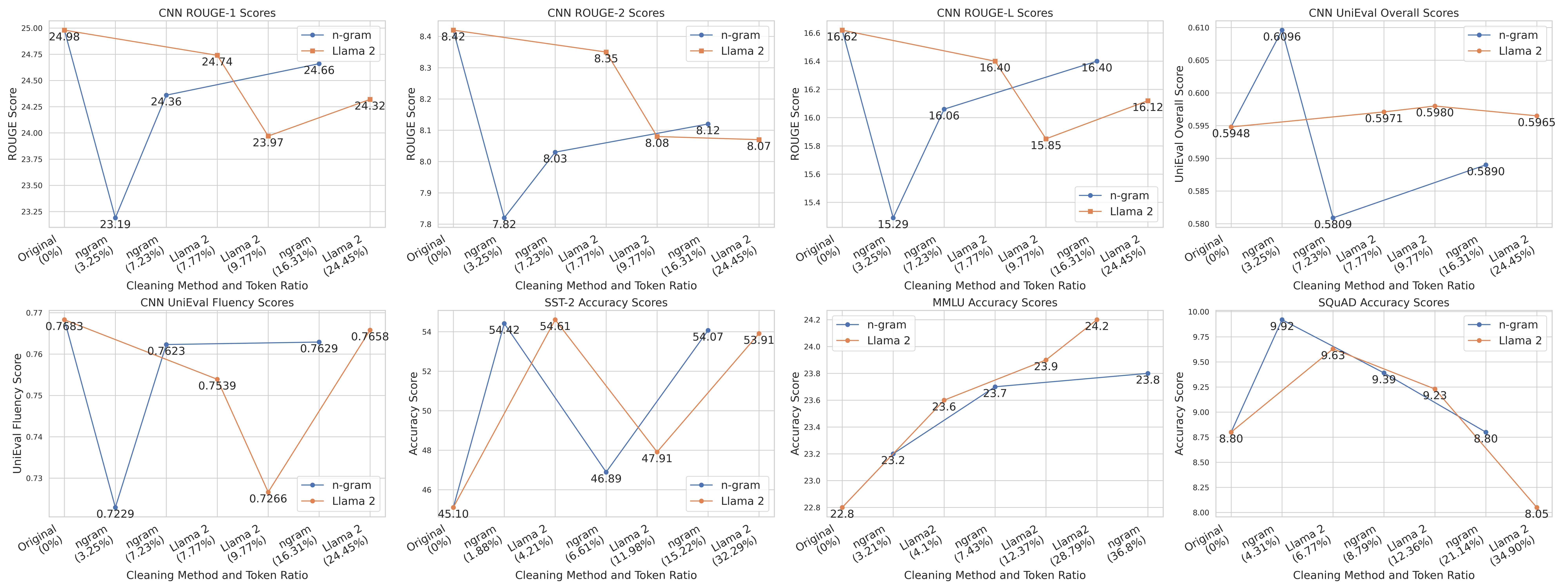}
    \caption{\textbf{Evaluation results on removing contamination from the pre-training corpus}. We deliberately select the parameters to achieve different ratios of removed tokens. The x-axis denotes the cleaning method (n-gram or Llama 2) followed by the percentage of tokens removed.}
    \label{fig::clean}
\end{figure}
In this section, we conduct experiments to clean the pre-training corpus based on the outlined n-gram and Llama 2 definitions. Specifically, the investigation aims to understand how the contaminated documents under these definitions would affect the performance if we filter them out of the pre-training corpus.
As described in \S\ref{sec::def}, we adopt different n-gram values $n$ for the direct n-gram overlap and Llama 2 contamination definitions, and we try various threshold $\lambda$ for the contamination percentage under Llama 2's definition.
These definitions are then used to filter ``contaminated'' documents out of the pre-training corpus, where a document is considered contaminated if any sentence in this document is considered contaminated. The detailed results are listed in Figure \ref{fig::clean}.

In our experimental setup, we systematically filter out a range of approximately $3\%$ to over $20\%$ of tokens labeled as ``contaminated'' from the pre-training corpus, aiming to analyze the effects of the percentage of tokens removed on the model performance.
The results, however, do not show a uniform pattern across different proportions of token removal. Interestingly, in certain instances where token removal exceeded $30\%$, the model's performance remained comparable to that of the original model. This finding raises questions about the accuracy of n-gram-based definitions for pinpointing effective contamination. It appears that documents excluded based on n-gram and Llama 2's definitions are not always genuinely contaminated, which reveals the insufficiency of such definitions for identifying effective contamination in practice.

We did not include PaLM's definition in our experiments since we found this definition is so strict compared to the other two definitions that very few documents would be filtered out. More analyses of the definitions are provided in Appendix \ref{sec::quant}, where we also extensively analyze the effects of varying the parameters of these definitions.

\subsection{Scaling Up with a Larger Model}
\label{sec::large}
We expand the experiment framework by incorporating GPT-2-large as the base model in our experiment. The primary objective is to assess if the effects of data contamination observed in smaller-scale models would persist in larger models. Due to computation constraints, we focus on the experiments on CNN and MMLU datasets for the ground-truth contamination with a contamination factor of 60, which is used to match the ratio of contamination with GPT-2-small experiments with a contamination factor of 10. 
A deviation in our setup compared to previous experiments is that we set a fixed number of training steps as opposed to a single epoch over the pre-training set; this is such that the training follows the compute-optimal scaling law for the available number of tokens. 

Despite the larger scale of the pre-training corpus in GPT-2-large, the impact of ground-truth contamination is clear. This finding underscores the significant influence of data contamination, which may remain concerning even in a large pre-training corpus.
\begin{table}[h]
    \centering
    \caption{\textbf{Evaluation results of GPT-2-large on CNN And Daily Mail and MMLU datasets}.}
    \label{table::genres}
    \resizebox{\textwidth}{!}{
    \begin{tabular}{l|c|cccccccc|cr}
        \toprule
        \textbf{Model} & \textbf{Parameters} & \multicolumn{8}{c}{\textbf{CNN And Daily Mail}} & \textbf{MMLU}\\
        \midrule
        & & Rouge-1 & Rouge-2 & Rouge-L & Coherence & Consistency & Fluency & Relevance & Overall & Accuracy\\
        \midrule

        GPT-2-large$_{\textit{original}}$ & 774M  & 27.47 & 9.67 & 17.74 &  0.6311 & 0.6910 & 0.8376 & 0.5942 & 0.6885 & 22.9\\
        GPT-2-large$_{\textit{gt}}$ & 774M & 28.43 & 10.85 & 18.74 & 0.6593 & 0.7335 & 0.8468 & 0.6082 & 0.7117 & 23.9\\
        \midrule
        GPT-2-large & 774M & 29.97 & 10.92 & 19.77 & 0.7259 & 0.8253 & 0.8997 & 0.6942 & 0.7863 & 23.0\\
        \bottomrule
    \end{tabular}
    }
\end{table}

\subsection{Assessing Evaluation-Level Contamination Analysis}
\label{sec::5}
In this section, we follow recent LLM reports \cite{chowdhery2022palm, touvron2023llama2} to divide evaluation data into different categories to see what we can learn from contamination analysis on the evaluation level.
Specifically, we follow Llama 2's definitions and methods~\cite{touvron2023llama2} to divide the evaluation data into four categories (``Clean'', ``Not Clean'', ``Not Dirty'', and ``Dirty'') and evaluate the model on each category separately.

\begin{table}[ht]
    \centering
    \caption{\textbf{The evaluation results on dividing the evaluation dataset into different categories.} We follow Llama 2's contamination definition and the associated parameters \cite{touvron2023llama2} to split the evaluation data. The parameters are shown as $n$ and $\lambda$, where $n$ is the n-gram value and $\lambda$ is the dirty and clean threshold, respectively.}
    \label{table::eval}
    \resizebox{\textwidth}{!}{
    \begin{tabular}{l|c|c|c|c|c|c|c|cr}
        \toprule
        Datasets & Model & Subset Type & $n$ & $\lambda$ & \# of Data & Avg. Contam. \%  & Results \\
        \midrule
        & &  & & & & & Overall\\
        \midrule
        \multirow{8}{*}{CNN} & \multirow{4}{*}{GPT-2-small$_{\textit{original}}$} & Clean & 15 & 0.85, 0.75 & 704 & 72.54 & 0.5743\\
        & & Not Clean & &  & 10,786 & 82.11 & 0.5920\\
        & & Not Dirty &  & & 9,203 & 80.22 & 0.5898\\
        & & Dirty &  & & 2,287 & 86.80 & 0.5955\\
        & \multirow{4}{*}{GPT-2-small$_{\textit{gt}}$} & Clean & 15 & 0.85, 0.75 & 704 & 72.54 & 0.6495\\
        & & Not Clean &  & & 10,786& 82.11 & 0.6986\\
        & & Not Dirty &  & & 9,203& 80.22 & 0.6950\\
        & & Dirty & & & 2,287& 86.80 & 0.6978\\
        \midrule
        & &  & & & & & F1 Score\\
        \midrule
        \multirow{8}{*}{SQuAD} & \multirow{4}{*}{GPT-2-small$_{\textit{original}}$} & Clean & 9 & 0.9, 0.7 & 571 & 67.10& 9.09\\    
        &  & Not Clean & & & 9,999 & 81.14 & 9.61\\
        & & Not Dirty & & & 9,741 & 78.91 & 9.59\\
        & & Dirty &  & & 856 & 97.03 & 9.24\\
        & \multirow{4}{*}{GPT-2-small$_{\textit{gt}}$} & Clean & 9 & 0.9, 0.7 & 571& 67.10 & 9.92\\
        &  & Not Clean & &  & 9,999&  81.14 & 11.39\\
        & & Not Dirty & &  & 9,741 & 78.91 & 11.37\\
        & & Dirty & &  & 856 & 97.03 & 10.21\\
        \bottomrule
    \end{tabular}
    }
\end{table}

We adopt relatively high clean/dirty threshold values $\lambda$ in order to arrive at similar portions of data for each category compared to Llama 2. 
We observed that the number of samples in each category is very sensitive to the selected $\lambda$ values.

We select CNN and SQuAD datasets and divide them into four categories based on the definitions and parameters described in Table \ref{table::eval}. We evaluate both the original model and the ground-truth contamination version of the model to see if the contamination will make a difference. Table \ref{table::eval} shows that the performance for the four categories is similar to each other. Even though the ``clean'' category under ground-truth contamination exhibited marginally lower results compared to the other categories, there was no clear indication that the ``dirty'' category outperformed the non-dirty categories. The fact from the previous experiments that the performance of the evaluated models can be boosted by contamination shows that these models are not immune to contamination in the pre-training corpus.

These results suggest that it may be insufficient to conclude that models are insusceptible to contamination based on such categorical evaluations. This draws attention to the need for more rigorous methodologies to assess the robustness of LLMs against data contamination accurately.

\section{Related Work}
\textbf{Data Contamination Definition and Investigation.} The exploration of data contamination has been a consistent element in LLM reports, dating back to the initial discussions of the memorization problem in BERT \cite{devlin-etal-2019-bert}.
Recent LLM reports \cite{radford2019rewon, brown2020language, chowdhery2022palm, openai2023gpt4, touvron2023llama, touvron2023llama2} have delved deeper into how evaluation data may be duplicated within pre-training corpora. 
These studies typically analyze the robustness of models against data contamination through n-gram-based definitions; the analysis is also typically focused on the evaluation level as opposed to the pre-training level (recall \S\ref{sec::1}).
However, such definitions may not accurately detect real contamination, casting doubt on the definitive conclusions drawn from these studies.
Recent LLM studies also investigated the embedding-based contamination definitions. The contamination analysis explored in phi-1/1.5 \cite{gunasekar2023textbooks, li2023textbooks} involves n-gram-based and embedding and syntax-based definitions but only focuses on code data. These studies represent a preliminary investigation in understanding the role of data contamination in the pre-training corpus. Another recent work~\cite{yang2023rethinking} shows that the existing n-gram-based and embedding-based definitions can be easily evaded by applying simple paraphrasing of evaluation data, emphasizing the urgent necessity for proper definitions of contamination and reliable detection methods. 

\textbf{Data Contamination and Memorization.}
Memorization in neural networks has been a well-explored topic in machine learning. Previous work has studied how memorization connects to and differs from generalization~\cite{NEURIPS2018_fface838, magar2022data, feldman2020does}, analyzed memorization in language models~\cite{carlini2023quantifying, nasr2023scalable}, and studied how memorization connects to privacy~\cite{ippolito2023preventing} and data extraction attacks~\cite{carlini2021extracting, nasr2023scalable}. 
Memorization is closely linked to data contamination as the model performance on evaluation data is no longer trustworthy if the evaluation data were memorized, regurgitated, and reasoned upon. 
Because of this connection, past work also explored membership inference attacks (MIA) for language models~\cite{mahloujifar2021membership, jagannatha2021membership, mireshghallah2022quantifying, carlini2022membership, mattern-etal-2023-membership, shi2023detecting}. However, these methods can sometimes be computationally intensive, and more generally, example-based matching can lead to false negatives in flagging contamination (\textit{e.g.} detection can be evaded through paraphrasing~\cite{yang2023rethinking}).
Other recent work has sought to identify pre-training data contamination heuristically by examining the likelihoods of texts after changing their ordering~\cite{oren2023proving} and of least probable tokens~\cite{shi2023detecting}.
Nevertheless, these methods are similarly inadequate for detecting textual transformations (\textit{e.g.} paraphrasing) and the heuristic nature of these methods may limit them from providing a clear understanding of how data contamination impacts the model performance on the pre-training level, highlighting a need for more comprehensive methods in this area of research.  

\section{Conclusion}

In this work, we conduct a \textit{pre-training level} analysis for the effects of data contamination on language models. 
We pre-train a series of GPT-2 models \textit{from scratch} to study the performance difference in different scenarios, underscoring the vital yet often overlooked role of ground truth in the context of data contamination detection.
This aspect is notably absent in existing studies. Our study also sheds light on the effects of repeated contamination on the performance of language models in downstream applications. 
Moreover, we critically assess the current n-gram-based contamination definitions as reported in recent LLM reports, revealing their inadequacy in accurately identifying true contamination within pre-training corpora.
Our replication of the existing robustness evaluations, which focus on evaluation level analysis that divides downstream datasets into different categories, suggests that such assessments fall short of affirming models' robustness to data contamination. 
Our findings highlight the need for more precise and effective contamination definitions, and the implementation of more stringent methods to ascertain the robustness of LLMs to data contamination. 

\section{Acknowledgements}

Research was supported in part by US DARPA KAIROS Program No. FA8750-19-2-1004 and INCAS Program No. HR001121C0165, National Science Foundation IIS-19-56151, and the Molecule Maker Lab Institute: An AI Research Institutes program supported by NSF under Award No. 2019897, and the Institute for Geospatial Understanding through an Integrative Discovery Environment (I-GUIDE) by NSF under Award No. 2118329. This work is also partially supported by NSF III 2046795, IIS 1909577, CCF 1934986, NIH 1R01MH116226-01A, NIFA award 2020-67021-32799, the Alfred P. Sloan Foundation, and Google Inc. Any opinions, findings, and conclusions or recommendations expressed herein are those of the authors and do not necessarily represent the views, either expressed or implied, of DARPA or the U.S. Government.
\clearpage
\bibliographystyle{plain}
\bibliography{neurips_2023}

\clearpage
\appendix
\section{Training Hyperparameters}
\label{sec::param}
We specify the hyperparameters we use for experiments for reproducibility and consistency of the results. In the GPT-2-small experiments, we set the \texttt{batch\_size=32, learning\_rate=0.0005, warmup\_ratio=0.01, weight\_decay=0.1}, and all other hyperparameters the same as the default settings of GPT-2-small. For the three runs of the main experiments, we adopt the seed numbers with \texttt{42, 1234, 2023} to ensure a fair comparison and consistency. For the GPT-2-large experiments, we set \texttt{batch\_size=128, learning\_rate=0.0001 random\_seed=42} instead to ensure training stability and keep all other parameters the same.

\section{Evaluation of Classification Tasks}
\label{sec::eval}
In this section, we describe the details of different prompts we utilized for the evaluation of SST-2 datasets. We select the prompts with different meanings and lengths to ensure the diversity of prompt formats, and the results for GPT-2$_{\textit{original}}$ are shown in Table \ref{table::sst2}. We can observe from the table that GPT-2-small models are quite sensitive to how prompts are structured in downstream tasks. This suggests we need more research to better understand and evaluate small language models on classification tasks, especially when the answers of the models are not within the label space, which can be addressed in future studies. 

\newcolumntype{M}[1]{>{\centering\arraybackslash}m{#1}}

\begin{table}[h]
    \centering
    \caption{SST-2 Accuracy Scores for the 10 Different Prompts.}
    \label{table::sst2}
    \resizebox{\textwidth}{!}{
    \begin{tabular}{M{6cm}|M{2cm}|M{2cm}|M{2cm}|M{2cm}|M{2cm}|M{2.2cm}|M{2cm}}
        \toprule
        \Large Prompts &  GPT-2-small$_{original}$& GPT-2-small$_{text}$ & GPT-2-small$_{gt}$& GPT-2-small & GPT-2-medium & GPT-2-large\\
        \midrule
       \Large Datasets   & SST-2 &   SST-2 &  SST-2 & SST-2 & SST-2 & SST-2\\
       \midrule
       \{text\} It is \{label\}  & 43.87 & 49.97 & 55.44 & 56.09 & 61.77 & 51.94\\
       \midrule
        \{text\} The text is \{label\}  & 42.98  & 49.81 & 55.60 & 54.36 & 58.47 &  53.92\\
        \midrule
        \{text\} The sentiment for this text is \{label\} & 44.20 & 48.27 & 51.73 & 51.55 & 54.38 & 45.83\\
        \midrule
        \{text\} The preceding text is \{label\} & 44.07 & 44.96 &  50.76 & 47.73 & 45.65 &  54.35\\
        \midrule
        \{text\} If the preceding text could be categorized as positive or negative, it would be \{label\} & 43.96 & 46.12 & 46.41 & 56.21 & 52.12 & 57.16\\
        \midrule
        \{text\} The sentence is \{label\} & 43.56 & 50.32 & 56.62 & 55.52 & 58.94 &  55.77\\
        \midrule
        \{text\} This text is \{label\} & 44.13 & 48.62 & 48.60 & 45.06 & 52.91 & 55.57 \\
        \midrule
        \{text\} Determine the sentiment of the preceding text: positive or negative: \{label\}  & 44.47 & 44.52 & 53.05 & 55.78 & 55.73&  55.85\\
        \midrule
        \{text\} The text belongs to \{label\} & 43.56 & 57.52 & 47.88 & 50.46 & 55.91 & 56.30 \\
        \midrule
        \{text\} The sentiment for this sentence should be \{label\} & 44.23 & 57.58 & 53.69 & 47.78 & 56.23 & 53.48\\
        \bottomrule
    \end{tabular}
    }
\end{table}

\section{More Discussions on Data Contamination}
\label{app::factor}
In this section, we show the experiment results for the AG News dataset, where we observe that the data contamination does not match the observation we had in our main experiments. 

\begin{figure}[ht]
    \includegraphics[width=\textwidth]{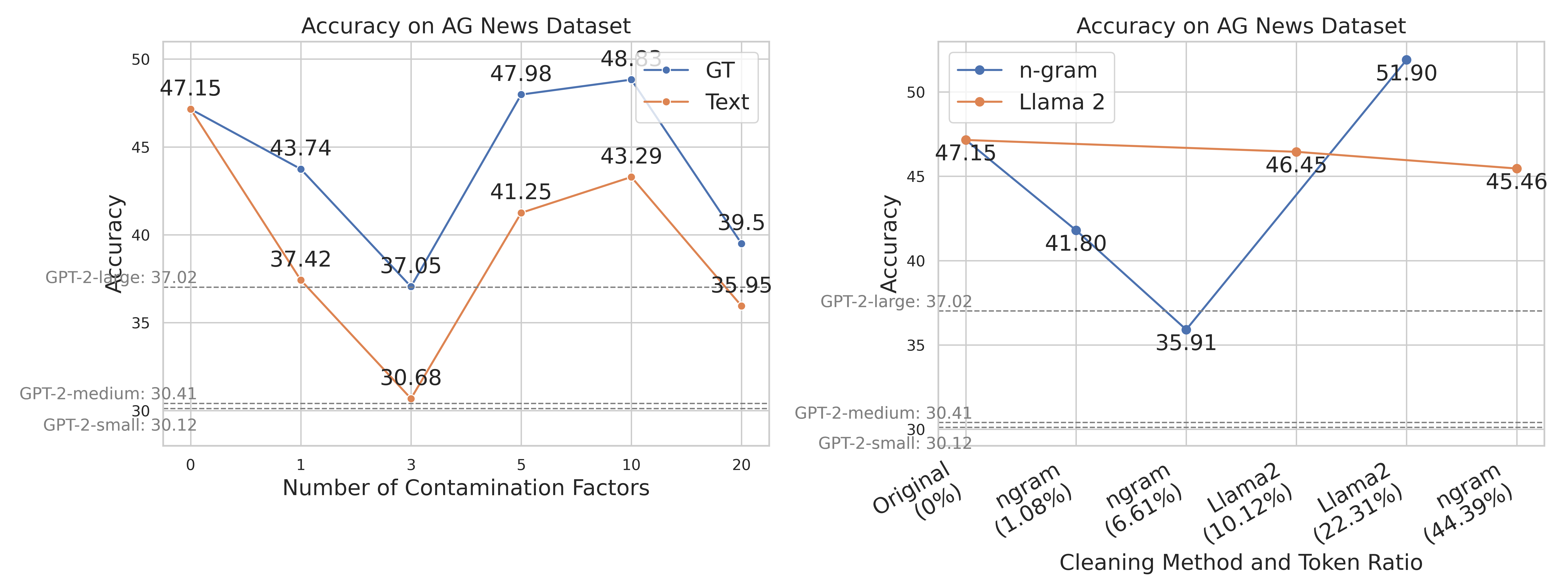}
    \caption{The evaluation results for AG News dataset on both contamination factor and removing contaminated data experiments. The performances for public model checkpoints from OpenAI are displayed as dotted lines in both figures.}
    \label{fig::agnews}
\end{figure}

We can observe that even the performance of the model pre-trained on the subsampled corpus is already higher than the OpenAI's public checkpoints. Interestingly, unlike previous experiments, we found that introducing text and ground-truth contamination does not significantly enhance performance. As we increase the contamination factors, the performance generally begins to decline at higher levels of contamination, as the U-shape trend in the previous experiment suggested, but with the lowest performance occurring at a contamination factor of 3. On the other hand, no matter how we increase the contamination factors, the performance is still much higher than the public checkpoints. One plausible explanation for this phenomenon is that the models may be assimilating or memorizing information from the AG News dataset present in the subsampled corpus. Consequently, the addition of various types of contamination does not yield substantial performance improvements and results in strange observations in this case.

This result suggests that the effects of data contamination on language models still require more effort to understand how knowledge of language models is constructed during pre-training.

\section{Quantitative Analysis for Contamination Definitions}
\label{sec::quant}
In this section, we analyze the different sets of parameters for different contamination definitions proposed in the previous studies to examine our evaluation dataset and pre-training corpus. We use the contamination ratio of the pre-training corpus for each evaluation dataset as a comparison to assess how strict these definitions are and the appropriate contamination definitions.
\subsection{N-Grams Direct Overlap}

\begin{figure}[ht]
    \includegraphics[width=\textwidth]{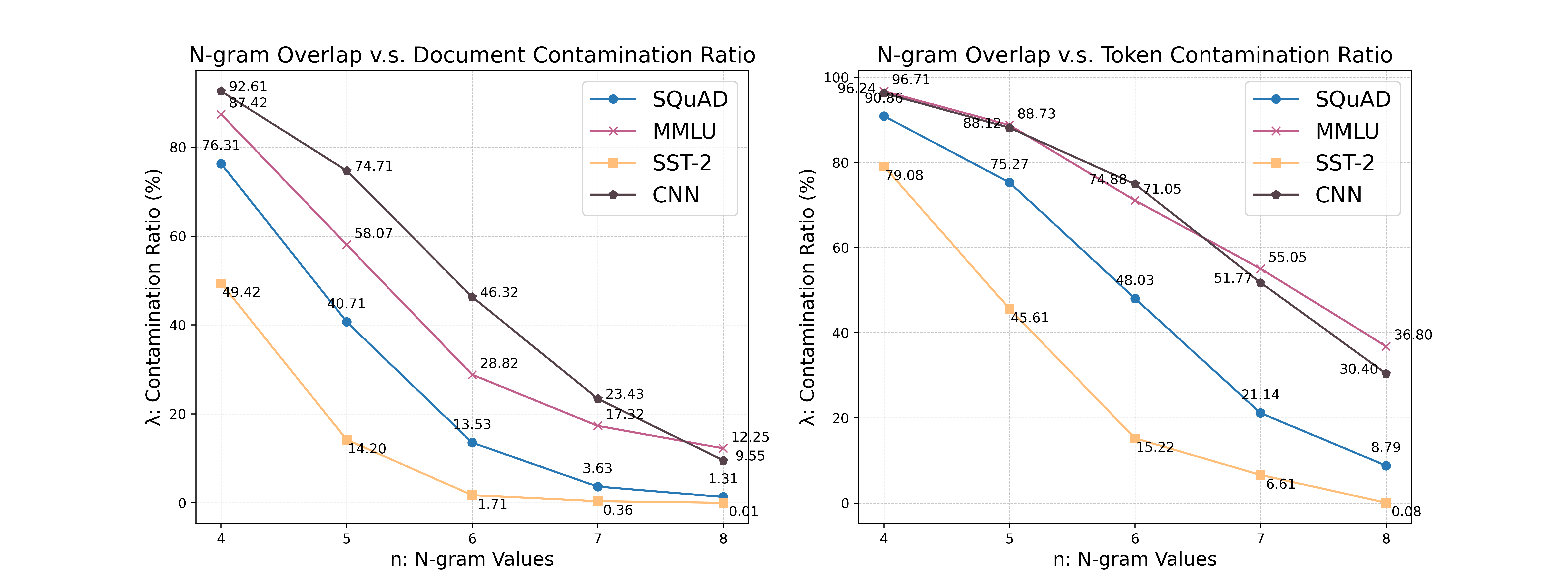}
    \caption{N-gram direct overlap contamination ratio w.r.t. different n-gram values for each dataset.}
    \label{fig::ngram}
\end{figure}
    
First, we examine the straightforward definition of contamination: the direct n-gram overlap within sentences of a training document. A training document is considered \textit{contaminated} if any $n$-gram in the document appears in the evaluation dataset. While this approach offers a direct measure of dataset duplication, its scope is limited. Solely relying on $n$-gram overlaps may overlook other forms of contamination since sentences can be rephrased in various ways, conveying identical meanings without any overlapping n-grams. Therefore, direct n-gram overlap is only to demonstrate how much of the content in the evaluation dataset appears in the pre-training corpus. During our filtering, a sentence is considered contaminated if any n-gram in the sentence appears in both pre-training data and evaluation data, and a document is considered contaminated if any sentence in this document is contaminated. We also report the total number of tokens in these documents that are considered contaminated. As shown in Figure \ref{fig::ngram}, we calculate the contaminated ratio of documents and tokens in the pre-training data for different $n$'s for comparison. We can observe that the contamination ratio varies for each dataset and how to define a reasonable threshold $n$ for the $n$-gram would be dependent on the text length of the evaluation dataset. For instance, in the SST-2 dataset, where many sentences comprise fewer than eight words, applying an 8-gram threshold would be impractical. Conversely, a very small n-gram value may fail to capture semantically meaningful content within sentences.

\begin{figure}[ht]
    \includegraphics[width=\textwidth]{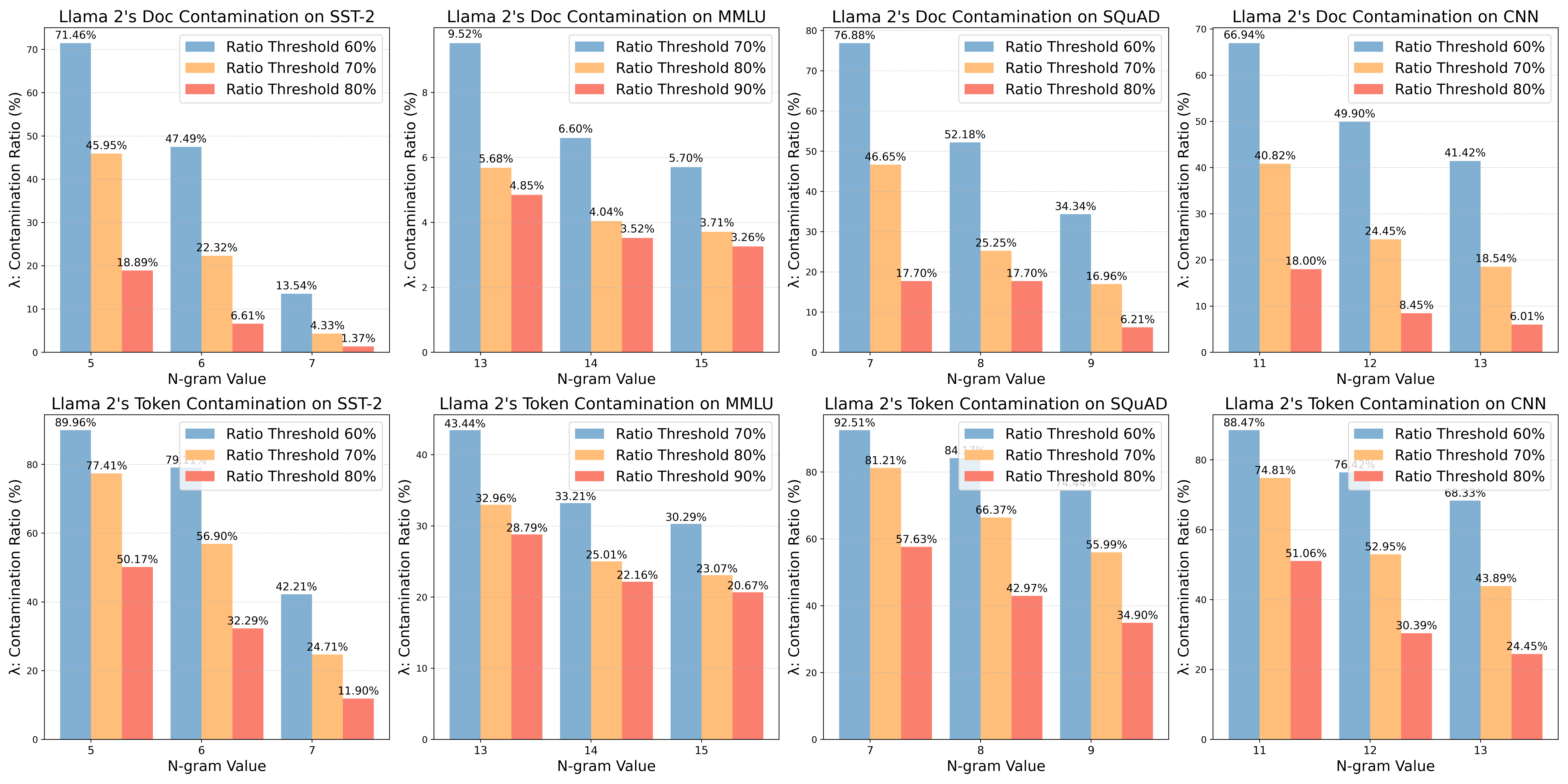}
    \caption{Contamination ratio for pre-training data based on Llama 2's definitions. We adopt the n-gram values that make the contamination ratio within a similar range and threshold from $60\% - 90\%$} for comparison.
    \label{fig:llama2}
\end{figure}

\subsection{PaLM and Llama 2's Definitions}
We conduct similar analyses for PaLM and Llama 2's definitions by considering different n-gram values $n$ and contamination threshold $\lambda$. PaLM's definition extends n-gram direct overlap to consider the overlapping percentage of n-grams in one sentence: a training document is considered \textit{contaminated} if more than $\lambda$ percentage of $n$-grams in a sentence of the document appear in the evaluation dataset. We observe that this definition is so strict that very few documents can satisfy it even if we relax $n$ and $\lambda$ to very small values compared to the original definition. The results for Llama 2's definitions are shown in Figure \ref{fig:llama2}. We report the percentage of contaminated documents and the percentage of tokens respectively. We can observe that the Llama 2 definitions lead to varied levels of identified contaminated documents and tokens, depending on the chosen parameters. These definitions concentrate on token contamination through n-gram duplication, which can be problematic because tokens may have different meanings in different contexts. Relying only on token duplication can misclassify sentences as contaminated. Additionally, similar to the straightforward n-gram definitions, setting the correct n-gram values and thresholds for different datasets remains a challenge with this approach.

We provide more detailed results for different parameters of these definitions, along with the PaLM's results, in Table \ref{table::palm} to better observe the trends for each definition.
\begin{table}[ht]
    \centering
    \caption{More results for the ratio of contaminated documents for different datasets with different definitions under different parameters.}
    \label{table::palm}
    \resizebox{\textwidth}{!}{
    \begin{tabular}{l|c|c|c|cr}
        \toprule
        Datasets & Filtering Method & N-gram Value & Threshold & \% of Contaminated Documents \\
        \midrule
        \multirow{4}{*}{SST-2} & PaLM & 5 & 10\% & $1.22\%$ \\
        & PaLM & 5 & 50\% & $\sim 0\%$\\
        & PaLM & 7 & 50\% & $\sim 0\%$\\
        & PaLM & 7 & 70\% & $0.0003\% $\\
        \midrule
        \multirow{6}{*}{SQuAD} & PaLM & 5 & 50\% & $0.077745\% $ \\
        & PaLM & 5 & 70\% & $0.007744\% $ \\
        & PaLM & 4 & 50\% & $0.081334\% $ \\
        & PaLM & 4 & 70\% & $0.037795\% $ \\
        & Llama 2 & 6 & 70\% & $76.38\% $ \\
        & Llama 2 & 6 & 80\% & $43.08\% $ \\
        \midrule
        \multirow{6}{*}{CNN} & PaLM & 7 & 70\% & $0.0896\% $ \\
        & PaLM & 8 & 70\% & $0.0302\% $ \\
        & Llama 2 & 14 & 70\% & $14.71\% $ \\
        & N-gram & 9 & - & $3.48\% $ \\
        & N-gram & 10 & - & $1.32\% $ \\
        & N-gram & 11 & - & $0.54\% $ \\
        \midrule
        \multirow{5}{*}{MMLU} & Llama 2 & 12 & 80\% & $6.76\% $ \\
        & Llama 2 & 15 & 95\% & $3.07\% $\\
        & Llama 2 & 18 & 90\% & $2.36\% $ \\
        & Llama 2 & 20 & 90\% & $1.92\% $ \\
        & Llama 2 & 24 & 90\% & $0.61\% $ \\
        \bottomrule
    \end{tabular}
    }
\end{table}

\end{document}